\RequirePackage{fix-cm}
\documentclass[smallextended]{svjour3}       
\smartqed  
\usepackage{natbib}
\usepackage{graphicx}
\usepackage{verbatim}
\usepackage{subfigure}
\usepackage{siunitx}
\usepackage{booktabs}
\usepackage{bbm} 
\usepackage{url} 
\usepackage[normalem]{ulem} 
\usepackage[dvipsnames]{xcolor} 
\usepackage[dvipsnames]{xcolor}

\newcommand{\edit}[2]{#2}
\journalname{Data Mining and Knowledge Discovery}
\begin{document}

\title{Visualizing Image Content to Explain Novel Image Discovery}



\author{Jake H.~Lee         \and
        Kiri L.~Wagstaff
}


\institute{Jake H.~Lee \at
     Columbia University \\
     \email{jake.h.lee@columbia.edu}           
     \and
     Kiri L.~Wagstaff \at
     Jet Propulsion Laboratory, California Institute of Technology\\
     4800 Oak Grove Drive, Pasadena, CA 91109 \\
     Tel.: +1-818-393-6393\\
     \email{kiri.l.wagstaff@jpl.nasa.gov}
}

\date{Received: date / Accepted: date}

\maketitle

\begin{abstract}

The initial analysis of any large data set can be divided into two
phases: (1) the identification of common trends or patterns and (2)
the identification of anomalies or outliers that deviate from those
trends.  We focus on the goal of detecting observations with novel
content, which can alert us to artifacts in the data set or,
potentially, the discovery of previously unknown phenomena.  To aid in
interpreting and diagnosing the novel aspect of these selected
observations, we recommend the use of novelty detection methods that
generate explanations.  In the context of large image data sets, these
explanations should highlight what aspect of a given image is new
(color, shape, texture, content) in a human-comprehensible form.  We
propose DEMUD-VIS, the first method for providing visual explanations of novel
image content by employing a convolutional neural network (CNN) to
extract image features, a method that uses reconstruction error to
detect novel content, 
and an up-convolutional network to convert CNN feature representations
back into image space.  We demonstrate this approach on diverse
images from ImageNet, freshwater streams, and the surface of Mars.
\edit{}{Finally, we evaluate the utility of the visual explanations
with a user study.}
\keywords{novelty detection, explanations, image analysis}

\end{abstract}

\section{Introduction}
\label{sec:intro}


Increases in computational power and data analysis capabilities have
inspired a corresponding increase in the appetite for data
collection.  Data sets collected by scientific, industrial, financial,
and political efforts continue to grow in scale and complexity.
Comprehending the contents of these data sets becomes challenging as
the number of items increases to thousands, millions, or more.  Data
sets that consist of images can be particularly problematic: while
humans are very good at image understanding, no human can feasibly
scan through and comprehend a collection of millions of images.

Automated machine learning methods can organize and prioritize image
collections to make the best use of limited human attention and time.
Classification methods can identify members of known classes, enabling
humans to quickly zero in on images of interest.  Unsupervised methods
enable exploration of data sets where the classes may not yet be
known.  For example, clustering methods can identify common groups or
trends within the data set.  Our focus is on a complementary
discovery task: highlighting novelties or anomalies within the data
set.  

Methods that identify novel observations help focus attention on items
that merit 
closer human examination~\citep{chandola:anom09}.  These are items that
can potentially teach 
us something new about the subject of study, inspire policy changes,
or overturn an existing scientific theory~\citep{kuhn:science62}.
Alternatively, they could signal instrument data collection errors,
artifacts, or processing problems and inspire important corrections or
upgrades.  Both kinds of anomalies are valuable to identify.

Identifying the anomalies is a necessary first step, but without
further information, a human reviewer may not be able to determine
what makes a given item anomalous.  This is especially true when the
data set is too large for any individual to have reviewed completely.
Users need to know what properties of the item are anomalous.  For
images, these properties might include color, shape, texture,
location, etc.  To date, very few anomaly detection methods provide
explanations.  We employ the DEMUD algorithm~\citep{wagstaff:demud13},
which expresses explanations as residual vectors in the input feature
space.  When the 
representation is derived from a neural network, additional steps are
needed to convert this explanation into a form that is understandable
for the human user.

In this paper, we describe DEMUD-VIS, the first method to generate
human-comprehensible (visual) explanations for novel discoveries in
large image data sets.  This approach has three steps: (1) convert each
image into a feature vector that captures semantic content, (2) employ
anomaly detection to discover images with novel content, and (3)
visualize the novel content for each selection in a
human-interpretable fashion. 
We employ existing methods to address the first two steps.  The
primary contribution of this paper is the final step: generating a
visual explanation of novel image content.
We evaluated DEMUD-VIS on several large image data sets, including ImageNet,
images of insect species found in freshwater streams on Earth, and
images collected by the Mars Science Laboratory rover.  We also
conducted a user study to assess whether the explanations are
useful to humans in identifying and understanding novel image content.
Given the ability to explain selections automatically chosen from very
large image data sets, investigators in a variety of application areas
can explore and learn new aspects of their own image data sets.


\section{Related Work}
\label{sec:relwork}

Developing explainable machine learning methods, models, and decisions
is a growing area of interest within the research community.  The goal
is to improve the interface between humans and machine learning so
that we can comprehend how a given method works, what a given model
learned, or why a given decision was made.

Following \edit{}{Rudin}, we define an {\em
interpretable} model as one in which model decisions can be directly phrased
in human-understandable ways, such as by a conjunction of attribute tests in a
decision tree \edit{}{\citep{rudin:stop-explaining18}}.
%
We contrast interpretable models with {\em explainable} models, for
which post-hoc explanations can be generated to describe the model
using a language other than that employed by the model itself.
Such an explanation can also be viewed as an ``interpretation'' of a
model's decision~\citep{biran:expl-survey17}.

Most of the work on interpretable or explainable models has focused
on supervised learning 
methods~\citep{biran:expl-survey17}.  Approaches for explaining the
decisions made by a supervised method range from using inherently
interpretable models such as rule lists via
CORELS~\citep{angelino:corels18}
to constructing post-hoc locally approximate models to explain the
decisions of a more complex model, as with
LIME~\citep{ribeiro:lime16}.  
%
In addition, 
methods from feature selection~\citep{guyon:featsel03} can be used to
identify the
features with the greatest influence on the model in general or on
particular decisions specifically.  
The DARPA XAI (Explainable AI) program seeks to support work across
the spectrum that includes interpretable models, explainable
features, and model approximation~\citep{gunning:xai19}.

Image classifiers may generate a
saliency map~\citep{dabkowski:saliency17} to indicate which parts of an
input image most influenced the classification decision.  The salience
map can indicate relevant content, but it does not explain why or how
a decision was made~\citep{rudin:stop-explaining18}.
Significant effort has been invested in trying to understand the
concepts being learned by deep convolutional neural
networks~\citep{montavon:interp-NN-18}, e.g.~by visualizing the 
features learned at each layer~\citep{zeiler:viscnn14}, linking
internal node activations to human concepts~\citep{bau:dissect17}, or 
generating synthetic inputs to visualize class membership with
DeepVis~\citep{yosinski:deepvis15}.
To help understand the network's view (representation) of a given
input image, methods such as Deep Goggle~\citep{mahendran:deepgoggle15}
and DeePSiM~\citep{dosovitskiy:upconv16} ``invert'' the internal
feature vector into \edit{reconstructed}{synthesized} images.



Fewer methods currently exist for generating explanations for
unsupervised machine learning methods, which are most relevant for
supporting discovery.
%
The MMD-critic method identifies prototypes and ``criticisms''
(outliers) to summarize the contents of a data
set~\citep{kim:mmd-critic-16}.  In this approach, individual items
help to describe the data set by example.  However, no
explanations are provided for why each item was selected as a
prototype or a criticism, so human viewers must construct their own
understanding of the content.
%
Other approaches seek to identify a subset of features to explain why
a given item should be considered an
anomaly~\citep{micenkova:expl-subset-13,dang:expl-feat-14,duan:expl-subspace-15}. 
The X-PACS method identifies feature subspaces that contain groups of
anomalies~\citep{macha:explain-anom-18}. 
\citet{siddiqui:seqfeatexp19} proposed a method to
generate {\em sequential} feature-based explanations to help human
experts decide whether a 
particular item was anomalous or not.  Their method provides a ranked
list of feature-value pairs that are incrementally revealed until
the human expert reaches a satisfactory level of confidence.  Features
are revealed in an order that minimizes the item's marginal
likelihood, assuming for example that items were generated from a
Gaussian mixture model.  The
quality of the explanations was assessed in terms of parsimony (the
number of feature-value pairs the expert needed to see to reach a
confident conclusion).

The DEMUD novelty detection algorithm provides an explanation that
considers all feature values simultaneously~\citep{wagstaff:demud13}.
DEMUD incrementally constructs a growing model of the user's knowledge
(using Singular Value Decomposition) and selects items with large
reconstruction error (those most likely to represent new classes) for
the user to review.  A DEMUD explanation consists of the
residual vector (information in the observation that was not captured
by the current model) for each selection.  DEMUD was designed to be
inherently interpretable, since the explanations employ the same
features by which items are originally described.  However, when
working with images, the features may not be individually
interpretable for humans.  In that case, an additional post-hoc
explanation step is required.


We developed the first approach to generate visually meaningful
explanations for class discovery in image data sets and
published it in short form at the 2018 ICML Workshop on Human
Interpretability~\citep{wagstaff:interp-images18}.  This approach
combines a CNN-based representation of semantic image content with the
DEMUD algorithm for image selection and an up-convolutional
network~\citep{dosovitskiy:upconv16} to convert image feature vectors
back into the image domain for visualization.  The current paper
provides more detail on the concept, approach, and experimental
results.  It also describes three new advances: (1) a new 
visualization step to yield more meaningful explanations
(described in Section~\ref{sec:method}), (2) several additional
experiments (Section~\ref{sec:viz-results}), including a new
scientific data set (stonefly images), a comparison of the use of
different CNN layers for image content representation, and a comparison of
novelty detection for class discovery and within-class novelty, and
(3) the results of a large user study to determine the utility and
impact of the visual explanations (Section~\ref{sec:userstudy}).


\section{Methods}
\label{sec:method}

\begin{figure}
\centerline{\includegraphics[width=4.5in]{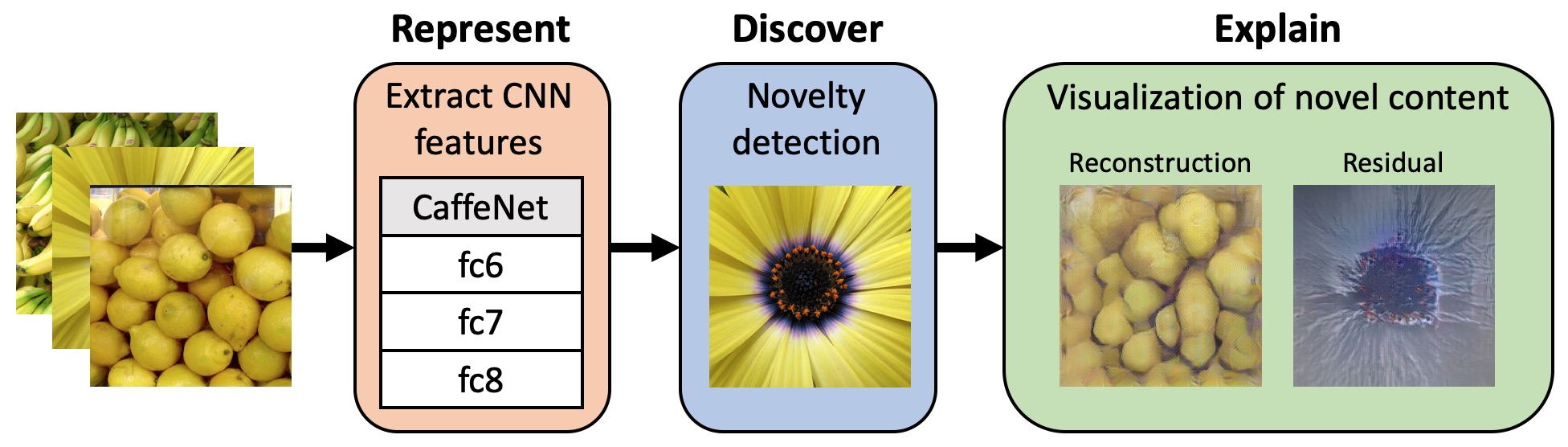}}
\caption{Novel image discovery with automatically generated visual
  explanations.  ``Reconstruction'' is the image content that the
  model recognized, and ``residual'' is the image content that was
  novel.} 
\label{fig:system}
\end{figure}

Our image discovery and explanation process consists of three steps:
\begin{enumerate}
	\item Represent image content as a feature vector.
	\item Discover images with novel content using an anomaly
          detection algorithm that also generates explanations.
	\item Render the explanations understandable to humans.
\end{enumerate}
To represent image content, we employ the representation extracted by a
convolutional neural network (CNN), as detailed in
section~\ref{sec:methodfeat}. We then rank images by their novelty
using a novelty detection algorithm that generates an explanation for
its selections (section~\ref{sec:methoddemud}). Finally, we
convert the algorithm's explanations into human-interpretable images
using CNN feature vector inversion (section~\ref{sec:methodvis}). 
Figure~\ref{fig:system} summarizes this process.  Here, the CNN is 
CaffeNet~\citep{jia:caffe14}, and the user can choose which of the
fully connected layers (fc6, fc7, fc8) to use for extracting image
content as a feature vector.  The yellow daisy image was identified as
a novel image, and the explanation indicates that the data model's
best reconstruction of that image was as a pile of yellow lemons.  The
color is close, but the spatial structure is wrong.  The residual
image shows the information that fell outside the model's
representation (i.e.,~the novel content in the daisy image); it 
emphasizes the dark center of the flower and the fine radial texture
in its petals.  The residual image is noticeably lacking in yellow,
because the color was already included in the reconstruction.

\subsection{CNN Features for Image Content Representation}
\label{sec:methodfeat}

Several methods from computer vision exist for extracting or
representing image content, 
such as LBP~\citep{ojala:lbp02}, SIFT~\citep{lowe:sift04}, and
HOG~\citep{dalal:hog05}. Recently, convolutional neural networks trained
on massive data sets have been shown to learn effective representations
of image content~\citep{donahue:decaf14,oquab:learning14}. Experiments
have shown that these representations can be used for a variety of
visual tasks, not just the task for which the original network was
trained~\citep{razavian:CNNfeat14}.

We extract a feature vector to represent each image by propagating the
image through a neural network that was previously trained for an object
classification task. Any network trained on a sufficiently diverse set
of inputs can be employed, although prior work has shown that supervised
networks provide better representations than unsupervised or
self-supervised networks~\citep{bau:dissect17}. For our experiments, we
used CaffeNet~\citep{jia:caffe14}, a CNN that consists of five
convolutional layers and three fully connected
layers~\citep{krizhevsky:alexnet12}. It was trained on 1.2 million
images from 1000 classes using the ImageNet data set~\citep{imagenet12}.


To extract image feature vectors, we resized each input image to 227
$\times$ 227 pixels, then propagated it through CaffeNet and recorded
the activations observed at a chosen 
fully connected layer (fc6, fc7, or fc8). 
\edit{}{These layers capture information at progressively more
  abstract levels.  For example, some spatial information is preserved
  in the fc6 representation, while less spatial information is
present in the fc7 layer, and progressively less in the fc8 layer,
which has a one-to-one mapping from neurons to classes.
The choice of which layer to use to represent image content can be
guided by knowledge of the kind of novelty that is meaningful in a
specific domain. 
}
For layers fc6 or fc7, we
recorded the activations prior to the linear rectification (ReLU)
activation function, which clips all negative values to zero
\citep{nair:relu10}. As we discuss in section~\ref{sec:methodvis}, the
pre-rectification activations carry information that is necessary for
the explanation generation process. Layers fc6 and fc7 each yield feature
vectors with 4096 values, while fc8, the final layer prior to the
softmax classification, yields a feature vector with 1000 values (one
per output class).

\subsection{Novelty Detection with Explanations}
\label{sec:methoddemud}

To detect novel images within a data set, we employed the DEMUD
method \citep{wagstaff:demud13}.
%
\edit{We chose DEMUD due to its demonstrated superiority to other
methods for class discovery and its unique capability to generate
per-selection explanations.}{DEMUD is a class discovery method that is
uniquely able to generate per-selection explanations. Other methods
such as SEDER \citep{he:seder09}, CLOVER \citep{huang:clover13}, and
RXD \citep{reed:rxd90} do not provide such explanations.}
DEMUD incrementally builds a model of what is known about a data set
$\mathbf{X}$, employing a singular value decomposition (SVD) to model the
most prominent data components.  DEMUD iteratively selects
the most interesting remaining item, with respect to the current SVD
model, and then updates the SVD model to incorporate (learn) the
newly selected item.  Interestingness (or novelty) is calculated using
reconstruction error, where a higher error indicates more novelty.
Reconstruction error $R$ for each item $\mathbf{x}$ is calculated as
\begin{equation}
R(\mathbf{x}) = ||\mathbf{x} - (\mathbf{U}\mathbf{U}^T (\mathbf{x} - 
\mu) + \mu) ||_2
\end{equation}
where $\mathbf{U}$ is the current set of top $k$ eigenvectors from the
SVD of $\mathbf{X_s}$, the set of already selected items, and $\mu$ is
the mean of all previously seen $\mathbf{x} \in \mathbf{X_s}$. The
number of principal components for the SVD model, $k$, is a
pre-specified parameter. The most interesting item $\mathbf{x}' =
\mbox{argmax}_{\mathbf{x} \in \mathbf{X}} \: R(\mathbf{x})$ is moved
from $\mathbf{X}$ to $\mathbf{X_s}$, and an incremental SVD
algorithm~\citep{lim:incremSVD05} updates $\mathbf{U}$ to incorporate
$\mathbf{x}'$.
This approach minimizes redundancy in the selections, since items
similar to any previously selected item will have low reconstruction
error. 

DEMUD's explanation for each selection $\mathbf{x}'$ is composed of
two parts. First, $\hat\mathbf{x}'$, the reconstruction of $
\mathbf{x}'$ after projection into the low-dimensional space defined by
$\mathbf{U}$, contains the information in $\mathbf{x'}$ that the
current model was able to represent:
\begin{equation}
\hat\mathbf{x}' =  \mathbf{U}\mathbf{U}^T(\mathbf{x}' - \mu) + \mu 
\end{equation}
Second, $\mathbf{r}$, the difference between $\mathbf{x'}$ and $\hat
\mathbf{x}'$ (referred to as the \textit{residual}), captures the
information in $\mathbf{x'}$ that the current model $\mathbf{U}$ was
\textit{not} 
able to represent:
\begin{equation}
	\mathbf{r} = \mathbf{x}' - \hat\mathbf{x}'
\end{equation}
Together, the reconstruction and residual can be examined to understand
which features and values led to each novelty selection.



\subsection{Visualization of Explanations}
\label{sec:methodvis}

When previous researchers applied DEMUD to numeric data sets, the
residual vectors $\mathbf{r}$ could be directly interpreted because
each feature was already represented as a human-comprehensible value
(e.g., size, age, or other measured property). \edit{In
contrast, in the CNN feature
domain, residual values for the 4096 features employed by layer fc6 in
CaffeNet are not directly interpretable.}{In contrast, residual vectors
in the CNN feature domain are not directly interpretable.}
Each feature value represents the activation of an individual neuron,
for which there may not always be a known, independent correspondence
to a particular image component. 
Prior work has proposed methods to determine an individual neuron's
sensitivity to specific image content through feature attribution or
example image generation~\citep{zeiler:viscnn14,olah:feature17}. While
these methods are useful for understanding the CNN model itself, they
are not sufficient to interpret a complete feature vector with 
thousands of neuron activation values.

To visualize DEMUD explanations in CNN feature space, we require a
method that can convert (or invert) CNN feature vectors back into the
image domain.
We considered several methods for this step.
Let $\Phi(x)$ be the feature vector the represents image $x$.
The Deep Goggle method employs gradient descent to generate a
synthetic input image $y$ that minimizes 
the Euclidean loss in {\em feature} space between $\Phi(x)$ and
$\Phi(y)$, the feature vector of the generated
image~\citep{mahendran:deepgoggle15}. 
\citet{dosovitskiy:upconv16} trained an up-convolutional (UC)
network to predict the original image $x$ given a feature vector
$\Phi(x)$ by minimizing the Euclidean loss in {\em image} space between
$x$ and $UC(\Phi(x))$, the output of the UC network. In later work, the
same authors replaced the pixel-space Euclidean loss with DeePSiM, a
weighted sum of feature-space loss, image-space loss, and an
adversarial loss determined by a discriminator
model~\citep{dosovitskiy:gen16}. While Deep Goggle is able to visualize
fine details such as texture, we found that colors and object locations
were often not faithful to the original image. On the other hand, the
UC method accurately represented colors and locations, but generated
blurry images; this is a well-known issue with generative models
optimizing image-space Euclidean loss~\citep{mathieu:videomse16}. The UC
DeePSiM method yielded the best results, generating feature vector
visualizations with sharp edges, detailed textures, and accurate shapes
and colors. \edit{We adopted the latter method for this study.}{We
adopted the latter method, which provided the implementation for
CaffeNet's fully connected layers.}

The UC DeePSiM model was trained to invert feature vectors that were
directly extracted from natural images, which differs from our
visualization goal.  We seek to visualize two synthetic items that
constitute the DEMUD explanation: the reconstruction
$\hat\mathbf{x}'$ and its residual $\mathbf{r}$. 
We made two changes to UC DeePSiM to enable its application to these
feature vectors. 

First, we modified the stage at which feature vectors are extracted 
for visualization.  \citeauthor{dosovitskiy:gen16} applied UC DeePSiM to the
feature vectors obtained after linear rectification (ReLU), which maps
all negative feature values to zero. However, the sign of the
values in the residual $\mathbf{r}$ expresses the relationship between
$\mathbf{x}'$ and $\hat\mathbf{x}'$ and, for our purposes, must be
preserved.  Therefore, we trained a new UC DeePSim network that operates
on the feature vectors obtained before rectification. 


Second, we found that UC DeePSiM sometimes failed to generate
meaningful images for the residual vectors $\mathbf{r}$.  Blank
or otherwise uninterpretable images were generated instead.  Recall
that $\mathbf{r}$ represents a partial image: the difference between
the original and reconstructed image vectors.  As a result, its
feature vector values can fall into a very different range than do those
obtained from natural images. 
Visualizations of fc7 residuals were successful, but those of fc6 and
fc8 sometimes failed.  While fc7's input and output are both fully
connected layers, fc6 is at the transition between convolutional and
fully connected layers, and fc8 is at the transition between fully
connected and output (softmax) layers.
\edit{To address the value range mismatch for fc6, we transformed the
residual vectors by shifting the mean $\mathbf{r}$ value to align with
the mean value of the reconstruction $\mathbf{x}'$}{ To improve the
visualization of fc6 residuals, we perform the following transformation
on each residual vector $\mathbf{r}$}:
\begin{equation}
\mathbf{r_{\text{shifted}}} = \mathbf{r} + \mathbbm{1} \cdot
(\bar{\mathbf{x}'}-\bar{\mathbf{r}}),
\end{equation}
where $\mathbbm{1}$ is a vector of ones, and $\bar{\mathbf{x}'}$ and
$\bar{\mathbf{r}}$ are the means \edit{values}{of the elements} of $
\mathbf{x}'$ and $\mathbf{r}$, respectively. \edit{}{This aligns the
mean of the residual's elements with the mean of the selected item's
elements.} This transformation
successfully yields meaningful visualizations of residual fc6 vectors.
However, it is not a complete solution for fc8 residuals, and further
improving fc8 visualizations is an ongoing area of investigation.

\section{Experimental Results}
\label{sec:results}

We conducted several experiments on benchmark and scientific data
sets to evaluate DEMUD-VIS in terms of (1) its ability to
discover new classes in progressively more difficult conditions and
(2) the quality of the generated explanations. 
All data sets, extracted features, and evaluation scripts are
available at \url{https://jakehlee.github.io/visualize-img-disc.html}. 

\subsection{Data Sets}
\label{sec:datasets}

\begin{figure}
  \centering
  \includegraphics[width=\textwidth]{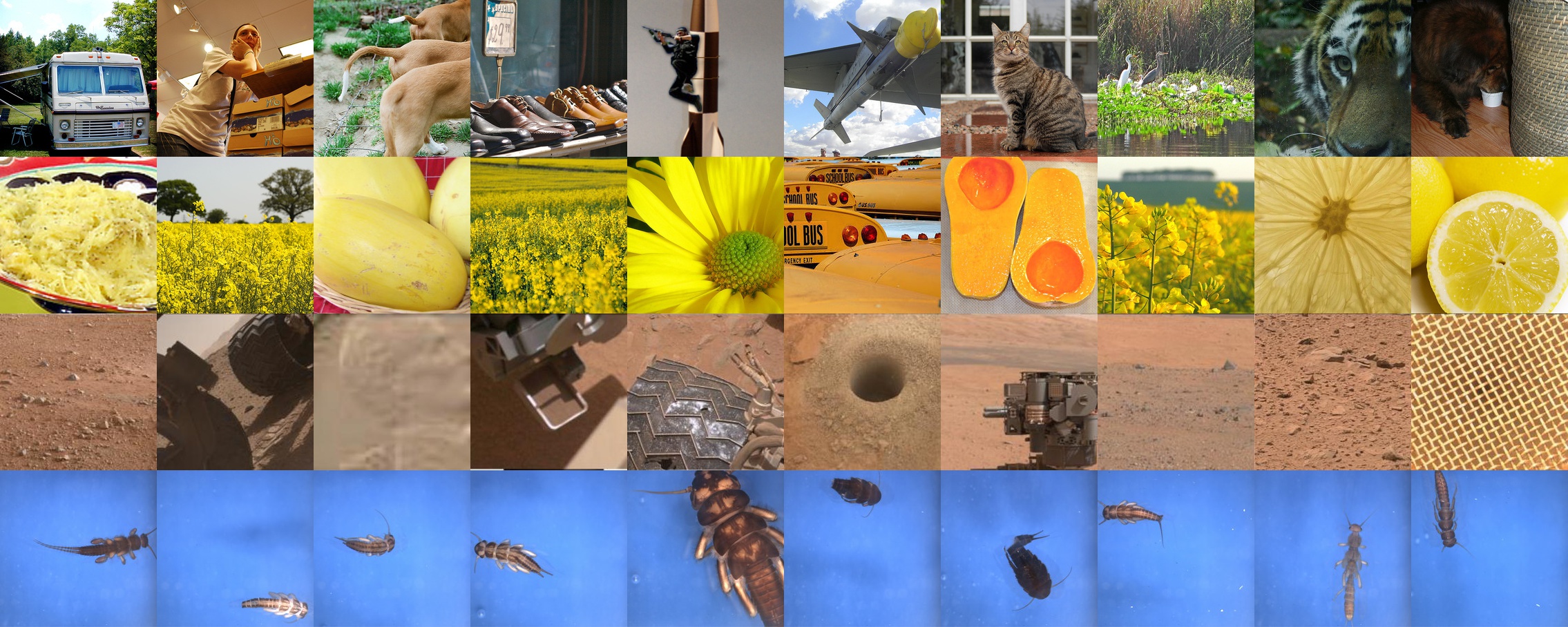}
  \caption{Randomly selected examples from each of the image data sets.
  From top to bottom:
  ImageNet-Random, ImageNet-Yellow, Mars-Curiosity, and STONEFLY9.}
  \label{fig:data}
\end{figure}

All images were center-cropped to enforce a square aspect ratio, then
resized to \edit{277x277}{227x227} pixels to match the input dimensions
of the CaffeNet CNN.

\paragraph{ImageNet-Random.} We
compiled a subset of the data set used by the ImageNet Large Scale
Visual Recognition Challenge in 2012 (ILSVRC2012)~\citep{imagenet12}.
We randomly selected 20 classes\footnote{Class definitions provided in
  code repository.}, 
then selected 50 images from each class in
the ILSVRC training set to create a heterogeneous yet \textit{balanced}
data set that contains 1000 images. The top row of
Figure~\ref{fig:data} shows ten random images from this data set. To
pose a more challenging task, we also created an \textit{imbalanced}
data set by selecting 50 images from the first 10 classes and only one
image from the last 10 classes for a total of 510 images.

\paragraph{ImageNet-Yellow.} For use in the
user study described in Section~\ref{sec:userstudy}, we created a
different ILSVRC subset designed to be more homogeneous in color.
We manually selected the following 10 classes as those containing
primarily yellow objects:
\texttt{banana}, \texttt{butternut squash}, \texttt{carbonara},
\texttt{corn}, \texttt{daisy}, \texttt{lemon}, \texttt{orange},
\texttt{rapeseed} (a yellow flower), \texttt{school bus}, and 
\texttt{spaghetti squash}.  By minimizing novelty in color, this data
set enabled us to test the detection, and explanation, of novelty that
depends on other kinds of differences such as shape, structure, and
semantic content. 
We selected up to 50 of the ``most yellow'' images from each class, which
are those images for which the mean pixel value had a Euclidean distance of
less than 150 from yellow (255, 255, 0) (except for the \texttt{school
bus} class, which used a cutoff of 170 to collect more images). The 
\texttt{daisy} and
\texttt{school bus} classes yielded only 19 and 26 images respectively,
while the other classes contained 50 images each, for a total of 445
images.  See row 2 of Figure~\ref{fig:data} for examples.

\paragraph{Mars-Curiosity.} We employed a publicly
available scientific data set composed of labeled images of the Mars
surface collected by the Mars Science Laboratory (Curiosity)
rover, plus an additional ``sun'' class\footnote{See code repository.}.
Identifying
images with novel content has an immediate operational value for Mars
planners, as new discoveries can influence plans for the rover's next
actions~\citep{kerner:novelty-iaai19}.  
This data set contains 6712 images from 25 classes collected by various
instruments on the Curiosity Mars rover from sol (Mars day) 3 to
1060~\citep{stanboli:mslv1,wagstaff:marsnet18}. The classes consist of broad
environmental categories such as ground and horizon, as well as
engineering categories such as instruments and calibration targets. The data
set is imbalanced, with class sizes ranging from 21 to 2684.
See row 3 of Figure~\ref{fig:data}.

\paragraph{STONEFLY9.} We also conducted experiments with a data set
that arises from the field of ecology.  The STONEFLY9 data
set\footnote{\url{http://web.engr.oregonstate.edu/~tgd/bugid/stonefly9/}}
consists of nine types of stoneflies that were collected in Oregon
streams and rivers, then imaged with a
microscope~\citep{lytle:stonefly10,martinez-munoz:stonefly09}.  In this
domain, the ability to automatically identify novel images is
beneficial for filtering out images of leaf fragments, debris, or
other insects not of interest to the study~\citep{lytle:stonefly10}.
We used the set0 subset, which contains 1362 images of 9 taxa (types)
of stonefly larva.  The data set is imbalanced, with class sizes
ranging from 44 to 223. All images were taken with a microscope on a
solid blue background. The classes are very highly similar, and
distinguishing between them is difficult even for trained human
experts~\citep{martinez-munoz:stonefly09}.
See row 4 of Figure~\ref{fig:data}.

\subsection{Experimental Methodology}

We conducted experiments in which we varied the novelty detection
method and the image content representation method.  This subsection
describes each variant and the metrics we employed.

\subsubsection{Novelty Detection Methods}

We compared DEMUD to a standard SVD and a random baseline. \edit{}{The
random baseline selects items from $\mathbf{X}$ randomly. This approach
yields no explanations for the selections, since no model is
constructed.  However, in many cases it can achieve good class
discovery results, especially for balanced data sets.} \edit{}{Other
rare category detection methods such as SEDER \citep{he:seder09},
CLOVER \citep{huang:clover13}, and RXD \citep{reed:rxd90} require a
labeling oracle and/or do not generate explanations.}

In all DEMUD and SVD experiments, $k$, the number of principal
components, was set to the maximum possible number of principal
components, which is the lesser of $d$ (dimensionality) and $n$
(number of items).  In many SVD applications, the goal is
dimensionality reduction, with $k$ much smaller than $d$.  In this
setting, our goal is to model as much of the information content as
possible so as to yield the best-quality reconstructions and therefore
be maximally sensitive to new image content.

SVD-based novelty detection is based on maximizing reconstruction
error, as with DEMUD:

\begin{equation}
R(\mathbf{x}) = ||\mathbf{x} - (\mathbf{U}\mathbf{U}^T (\mathbf{x} - 
\mu) + \mu) ||_2
\end{equation}

For an SVD, the model $\mathbf{U}$ and mean $\mu$ are constructed from
the entire data set $\mathbf{X}$, while for DEMUD, $\mathbf{U}$ and
$\mu$ represent only the prior selections in $\mathbf{X_s}$ and
therefore incrementally grow and change.  DEMUD requires the
specification of how to initialize $\mathbf{X_s}$.  In these
experiments, we initialized $\mathbf{X_s}$ by constructing a full SVD
over the entire data set and selecting the single item with the
highest reconstruction error. 

\edit{The random baseline selects items from $\mathbf{X}$ randomly. 
This approach yields no explanations for the selections, since no model is
constructed.  However, in many cases it can achieve good class discovery
results, especially for balanced data sets.}{}

\subsubsection{Image Content Representation Methods}

We assessed three options for representing image content as a feature
vector for input to the novelty detection methods.

\begin{enumerate}
  \item \textit{Pixel}: Pixel values of the image, flattened into a
    one-dimensional 
  vector by reading the image array (where the color channel is the
  last axis) in C-like order.
  \item \textit{SIFT}: Features extracted from Scale Invariant Feature
  Transform (SIFT) keypoints~\citep{lowe:sift04}.
  \item \textit{CNN}: Features extracted from the fully connected
  layers of CaffeNet, as described in Section~\ref{sec:methodfeat}.
\end{enumerate}

SIFT-based features were generated using a visual bag of words
approach. First, we used OpenCV's SIFT module to extract SIFT keypoints
from every image in the data set~\citep{bradski:opencv00}. We then ran
$k$-means clustering (with $k=k_{\text{SIFT}}$ clusters) on all
keypoints. Finally, for each image, we assigned each of its keypoints
to the nearest cluster to generate a histogram of cluster counts per
image. This resulted in a $k_{\text{SIFT}}$-dimensional feature vector
for each image, which can be used directly by DEMUD or a standard SVD.
Since there is no standard way to select the best $k_{\text{SIFT}}$ in
advance, we report the best performance achieved after testing
$k_{\text{SIFT}} = {3,4,5,10,15,20}$.

Note that SIFT-based representations do not provide interpretable
explanations.  The resulting reconstructions and residuals are in the
form of distributions of unexpected values for keypoint cluster
histograms and do not have a natural correspondence to the visual
domain.  As with the random baseline, SIFT is included in these
experiments as a point of comparison for class discovery performance;
it enables the first and second parts of the task (representation and
discovery) but not the third part (explanation).  

\subsubsection{Evaluation Metrics}

We assessed novelty detection in terms of the ability to discover new
classes.  While novelty detection is unsupervised and therefore does
not have access to class labels while operating, we used the labels to
quantify class discovery after the fact.  After each selection $i$, we
recorded $C_i$, the cumulative number of distinct classes discovered
up to and include that selection.  Plotting the 
number of discovered classes as a function of the number of items
selected yields a discovery curve. We calculated the area under this
curve for the first $t$ selections by summing $C_i$ from selection 1
to $t$. We then divided this value by the area under a perfect
discovery curve to calculate the normalized area under the curve
($nAUC_t$):
\begin{equation}
  nAUC_t= \frac{\sum_{i=1}^t C_i}{\left(\sum_{i=1}^c i \right) + c (t-c)} \times 100.0,
\end{equation}
where $c$ is the number of classes in the data set and the denominator
is the perfect discovery performance with a new class is discovered
in each of the first $c$ selections. 

The value $t$ (number of selections made) reflects the number of
selections that the user wants to review.  In practice, this parameter
will vary depending on $c$ (which may be unknown), the size of the
data set, and human time available.  For our experiments, we set $t$ to 
either the number of selections needed to discover all classes
via random selection, or $300$, whichever is less. 


To assess the class discovery performance of the random baseline, we
calculated the average $nAUC_t$ over 1000 trials.  DEMUD and SVD are
deterministic, so a single trial suffices to assess their output.

The second part of our evaluation was to assess the quality and utility
of the explanations.  Assessing explanations in an objective fashion
is notoriously challenging~\citep{montavon:interp-NN-18}, especially
for unsupervised settings in which there is no single correct
output~\citep{siddiqui:seqfeatexp19}. 
We evaluated the generated explanations through a user study, which 
is described in Section~\ref{sec:userstudy}.

\subsection{Class Discovery Performance}
\label{sec:disc-perf}


\subsubsection{ImageNet Data Sets}

\begin{figure}[]
\centering
\subfigure[Pixel, SIFT, and CNN representations]{
  \includegraphics[width=0.475\textwidth]
  {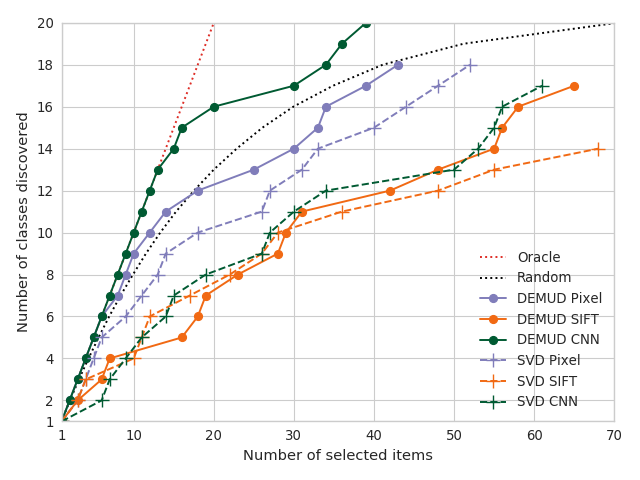}
}
\subfigure[CaffeNet layers fc6, fc7, and fc8]{
  \includegraphics[width=0.475\textwidth]
  {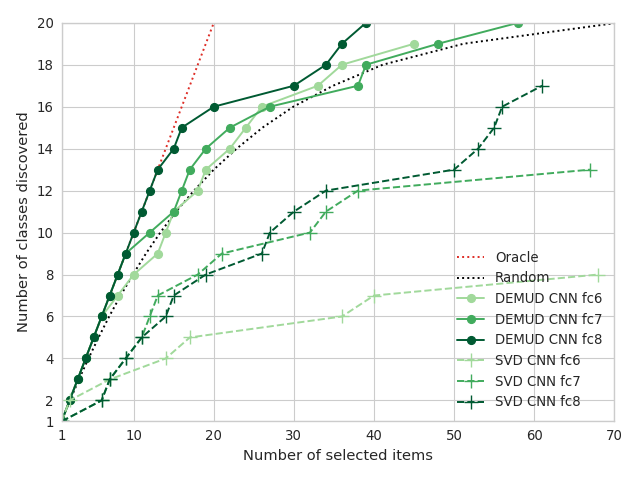}
}
\caption{\edit{Discovery of 20 balanced ImageNet-Random classes}
  {Discovery of 20 classes from balanced ImageNet-Random (50 images
  per randomly selected ImageNet class, 1000 images total)}, up to
  $t=70$ selections. DEMUD with CNN layer fc8 features achieved the
  best performance.} 
\label{fig:exp-bal}
\end{figure}

Our first experiment employs the ImageNet-Random data set, which
consists of 20 balanced classes.
Figure~\ref{fig:exp-bal}(a) shows the number of classes discovered as
a function of the number of selections.  
The ``Oracle'' line shows (hypothetical) perfect discovery performance. 
Since CaffeNet was optimized to distinguish between 1000 pre-defined
image classes that include the 20 classes in this data set, we
expected to see high performance when using CaffeNet feature vectors
(green lines).  However, we found that DEMUD using CNN-based
representations was the only method to out-perform random selection.
SVD using the CNN features performed much worse than random.
The same was true for DEMUD and SVD when using SIFT features or pixel
values.  However, DEMUD using CNN layer fc8, the last layer of
CaffeNet, achieved oracle-level performance for its first 13 selections
(discovered a new class with each selection).
Figure~\ref{fig:exp-bal}(b) compares DEMUD and SVD performance using
different CaffeNet layers to represent image content.  DEMUD using fc8
employs feature vectors that are most closely aligned with CaffeNet's
final classification output, so it is unsurprising that this layer
yielded the best performance.  However, DEMUD using fc6 and fc7
performed almost as well, suggesting that novel content is captured
even at lower levels of the network.  SVD's performance was more
variable, but again fc8 yielded the best result.

\begin{figure}[]
\centering
\subfigure[Pixel, SIFT, and CNN representations]{
  \includegraphics[width=0.475\textwidth]
  {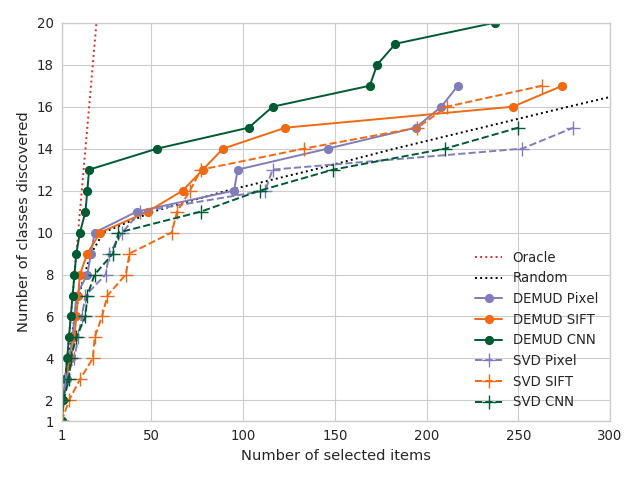}
}
\subfigure[CaffeNet layers fc6, fc7, and fc8]{
  \includegraphics[width=0.475\textwidth]
  {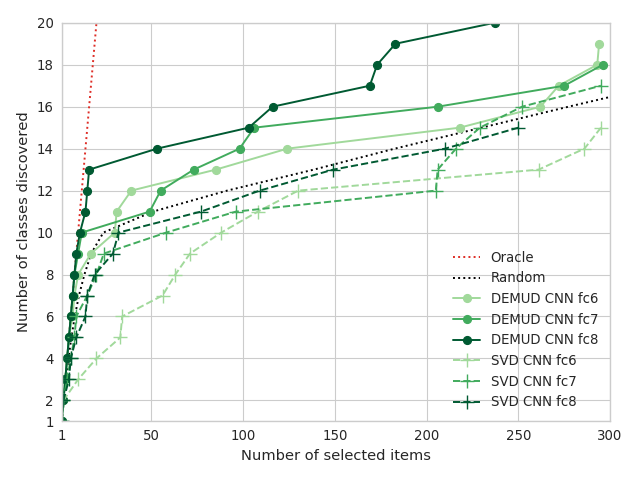}
}
\caption{\edit{Discovery of 20 imbalanced ImageNet-Random classes}
{Discovery of 20 classes from imbalanced ImageNet-Random (50 images
from 10 random classes, one image from 10 random classes, 510 images
total)}, up to $t=300$ selections. DEMUD with CNN layer fc8
features achieved the best performance.}
\label{fig:exp-ubal}
\end{figure}

For real discovery problems, classes are unlikely to be equally
balanced in the data set.  The imbalanced ImageNet-Random data set was
more challenging; while the random baseline discovered all 20 balanced
classes within 70 selections, it found only 16 after 300
selections in the imbalanced data set (Figure~\ref{fig:exp-ubal}(a)).
In fact, DEMUD using CNN layer fc8 was the only method to discover all 20
classes within the first 300 selections. DEMUD again out-performed SVD
for all representations, and DEMUD achieved better performance with
the CNN-based representations over other representations. 
Figure~\ref{fig:exp-ubal}(b) shows that DEMUD and SVD had more
similar performance on this data set, but the SVD remained below the
random baseline.

\begin{figure}[]
\centering
\subfigure[Pixel, SIFT, and CNN representations]{
  \includegraphics[width=0.475\textwidth]
  {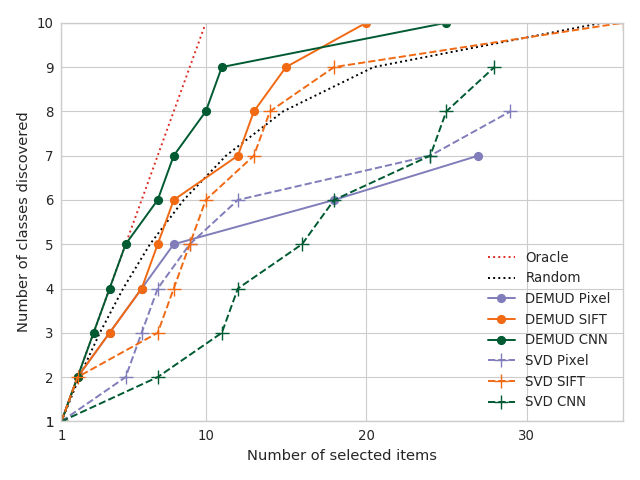}
}
\subfigure[CaffeNet layers fc6, fc7, and fc8]{
  \includegraphics[width=0.475\textwidth]
  {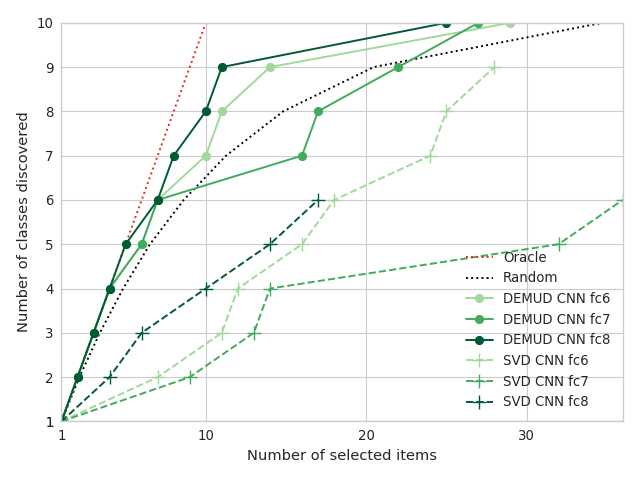}
}
\caption{\edit{Discovery of 10 ImageNet-Yellow classes}
{Discovery of 10 classes from ImageNet-Yellow (up to 50 images from
10 classes selected for color similarity, 445 images total)}, up to
$t=36$ selections. DEMUD with CNN layer fc8 features achieved the best
performance.}
\label{fig:exp-y10}
\end{figure}

Finally, Figure~\ref{fig:exp-y10}(a) shows results for the
ImageNet-Yellow data set. This discovery task is easier because there
are only 10 classes, not 20.  Once again, DEMUD using CNN features
achieved the best performance, fc8 yielded the best representation,
and fc6 and fc7 were not far behind (Figure~\ref{fig:exp-y10}(b)).

\begin{figure}[]
  \centering
  \subfigure[Selections 1 through 10.]{
    \includegraphics[width=\textwidth]{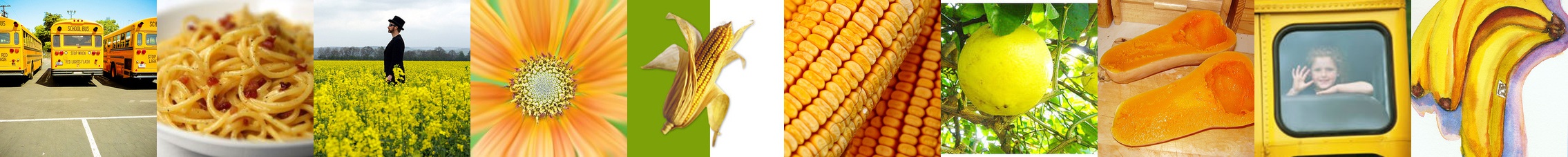}
  }
  \subfigure[Selections 436 through 445.]{
    \includegraphics[width=\textwidth]{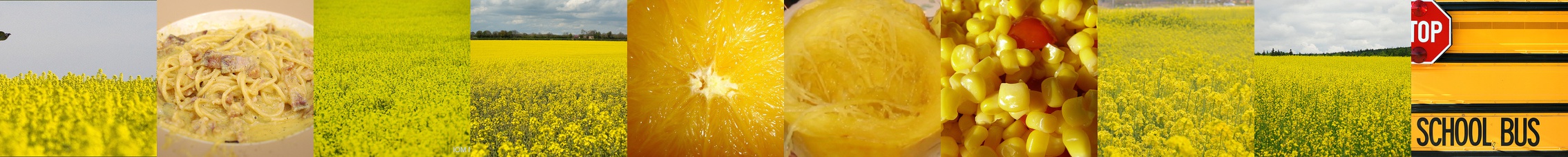}
  }
  \caption{The first and last ten ImageNet-Yellow images selected by
  DEMUD using CNN layer fc8 representations. Note the higher diversity
  of content in the first ten selections.}
  \label{fig:y10-topbot}
\end{figure}

Figure~\ref{fig:y10-topbot} shows the first ten and last ten images
obtained when using DEMUD with fc8 to select (and therefore rank) all
445 images in the ImageNet-Yellow data set. 
The first ten selections are
very diverse, comprising images from eight different classes.  The
classes that appear twice (\texttt{school bus} and \texttt{corn})
include images with very different scales and backgrounds.
The last ten selections (Figure~\ref{fig:y10-topbot}(b)) 
contain images with less variation in color,
texture, and backgrounds.

\begin{table}[]
\centering
\caption{Discovery $nAUC_{t}$ on ImageNet subsets (best
result for each data set in {\bf bold}).}
\label{tab:in-auc}
\begin{tabular}{@{}lllllll@{}}
\toprule
 & \multicolumn{2}{l}{Balanced ($nAUC_{70}$)} & 
 \multicolumn{2}{l}{Imbalanced ($nAUC_{300}$)} & 
 \multicolumn{2}{l}{Yellow ($nAUC_{36}$)} \\
Features & DEMUD & SVD & DEMUD & SVD & DEMUD & SVD \\ \midrule
CNN-fc6 & 86.36 & 31.65 & 69.09 & 52.62 & 89.84 & 60.63 \\
CNN-fc7 & 87.93 & 54.05 & 72.32 & 59.74 & 84.13 & 37.14 \\
CNN-fc8 & {\bf 94.13} & 60.58 & {\bf 84.08} & 61.76 & {\bf 93.33} &
53.97 \\
\midrule
SIFT$_{(k_{\text{SIFT}})}$  & $64.79_{(5)}$ & $53.97_{(5)}$ & $70.64_{
(5)}$ &
$66.52_{(4)}$ & $89.52_{(20)}$ & $80.00_{(20)}$ \\
Pixel & 81.16 & 73.80 & 69.38 & 61.53 & 61.27 & 64.44 \\ \midrule
Random & \multicolumn{2}{c}{83.06} & \multicolumn{2}{c}{63.46} & 
\multicolumn{2}{c}{79.19} \\
\bottomrule
\end{tabular}
\end{table}

Table \ref{tab:in-auc} compiles the $nAUC_{t}$ values for the
ImageNet-Random (both balanced and imbalanced variants) and
ImageNet-Yellow data sets. DEMUD using CNN layer fc8 was the best
performing method for all subsets of ImageNet. 

\subsubsection{Mars-Curiosity}

\begin{figure}[]
\centering
\subfigure[DEMUD and SVD performance across representations]{
  \includegraphics[width=0.475\textwidth]
  {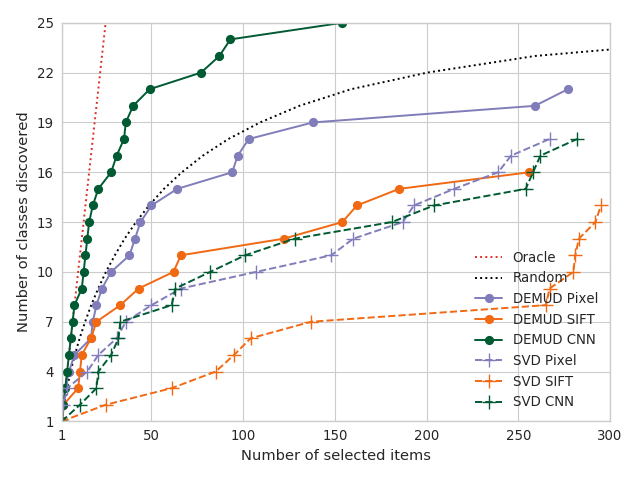}
}
\subfigure[DEMUD and SVD performance across CaffeNet layers]{
  \includegraphics[width=0.475\textwidth]
  {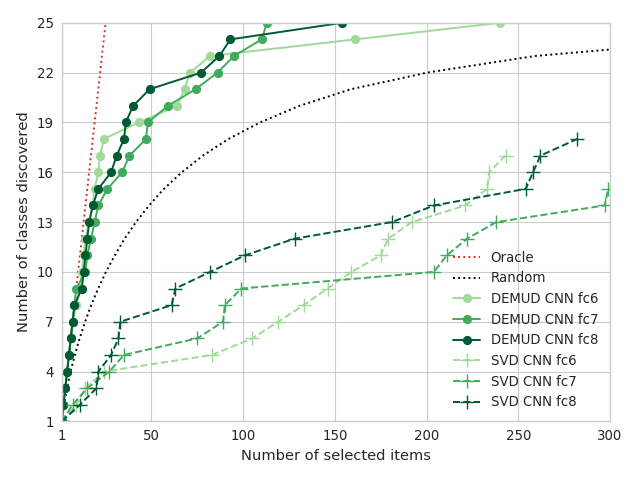}
}
\caption{\edit{Discovery of 25 Mars-Curiosity classes}
{Discovery of 25 classes from Mars-Curiosity (6712 total images, class
sizes ranging from 21 to 2684)}, up to $t=300$
selections. DEMUD with CNN layer fc8 features achieved the best
performance.}
\label{fig:exp-msl}
\end{figure}

\begin{table}[b]
\centering
\caption{Discovery $nAUC_{300}$ on Mars-Curiosity (best
result in {\bf bold}).}
\label{tab:msl-auc}
\begin{tabular}{@{}lll@{}}
\toprule
 & \multicolumn{2}{l}{Mars-Curiosity }\\ 
Features & DEMUD & SVD \\ \midrule
CNN-fc6 & 91.51 & 39.54 \\
CNN-fc7 & 92.56 & 36.19 \\
CNN-fc8 & {\bf 93.75} & 47.17 \\ \midrule
SIFT$_{(k_{\text{SIFT}})}$ & $50.83_{(20)}$ & $24.21_{(20)}$ \\
Pixel & 69.32 & 47.64 \\ \midrule
Random & \multicolumn{2}{c}{75.97} \\ \bottomrule
\end{tabular}
\end{table}

To assess whether novelty detection using CaffeNet feature vectors
could be effective in a different image domain, we applied 
the same techniques to the Mars-Curiosity data set.  
The reader may wonder which ImageNet classes the CaffeNet model
proposed for the Mars-Curiosity images.  We found that the most
commonly predicted classes were \texttt{sand viper} and
\texttt{nematode} for images of the Mars surface and visually similar
classes for the rover parts, such as \texttt{waffle iron} for the
rover wheels, which have a metal grill texture.
Since the network was not trained on Mars-specific classes, 
this experiment tests whether abstract representations learned by
CaffeNet from Earth images can generalize sufficiently well to the
Mars domain.

As shown in Figure~\ref{fig:exp-msl} and Table~\ref{tab:msl-auc}, this
more challenging data set generally yielded lower discovery
performance than the 
ImageNet data sets, except for DEMUD.  DEMUD with CNN-based
representations was again the only method to perform better than random
selection. It was also the only method to discover all 25 classes
within 300 selections. The fc8 layer yield the best performance, but
all three performed well, which is consistent with our results from
previous experiments.

\begin{figure}[]
  \centering
  \subfigure[Selections 1 through 10.]{
    \includegraphics[width=\textwidth]{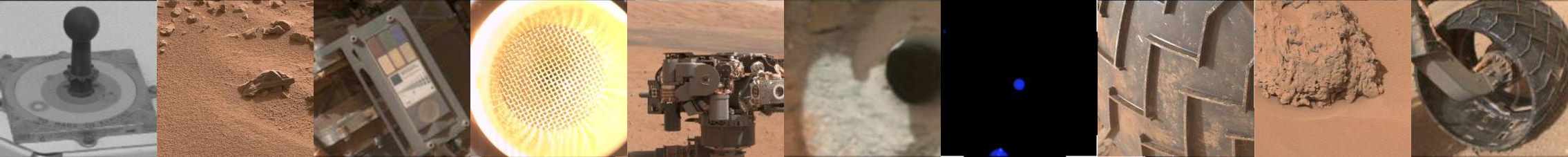}
  }
  \subfigure[Selections 6703 through 6712.]{
    \includegraphics[width=\textwidth]{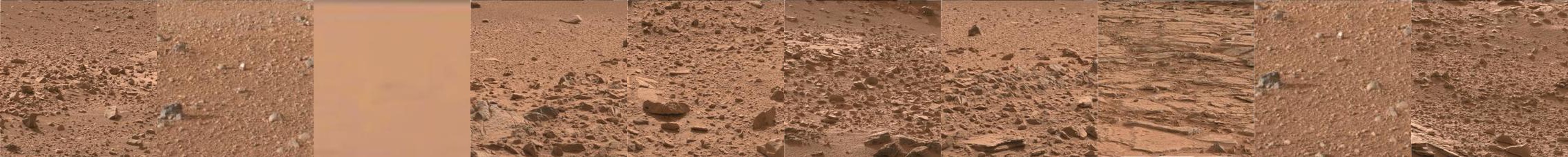}
  }
  \caption{The first and last ten Mars-Curiosity images selected by
  DEMUD using CNN layer fc8 representations. Note the higher diversity
  of content in the first ten selections.}
  \label{fig:msl-topbot}
\end{figure}

Figure~\ref{fig:msl-topbot} shows the first ten
and last ten selections made by this method. Once again, the first ten
selections are diverse in content, with eight different classes 
represented in these images. Even for repeat classes, such as the eight
and tenth images from the left (which are both in the \texttt{wheel} class),
the content itself differs significantly. Meanwhile, the last ten
selections made by the method are all from the same class,
\texttt{ground}, and exhibit minimal novelty.

\subsubsection{STONEFLY9}
\begin{figure}[b]
\centering
\subfigure[DEMUD and SVD performance across representations]{
  \includegraphics[width=0.475\textwidth]
  {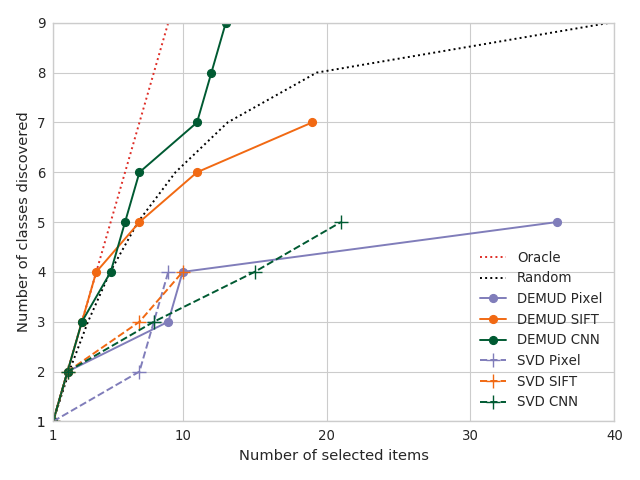}
}
\subfigure[DEMUD and SVD performance across CaffeNet layers]{
  \includegraphics[width=0.475\textwidth]
  {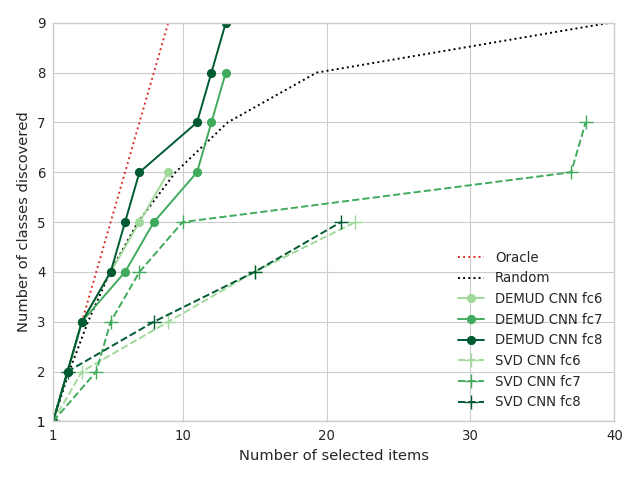}
}
\caption{\edit{Discovery of nine STONEFLY9 classes}{Discovery of nine
classes from STONEFLY9 (1362 images in highly similar classes, class
sizes ranging from 44 to 223)}, up to $t=40$ selections.
DEMUD with CNN layer fc8 features achieved the best performance.}
\label{fig:exp-s9}
\end{figure}

\begin{table}[]
\centering
\caption{Discovery $nAUC_{40}$ on STONEFLY9 (best
result in {\bf bold}).}
\label{tab:s9-auc}
\begin{tabular}{@{}lll@{}}
\toprule
 & \multicolumn{2}{l}{STONEFLY9} 
 \\
Features & DEMUD & SVD \\ \midrule
CNN-fc6 & 67.59 & 47.84 \\
CNN-fc7 & 83.95 & 57.10 \\
CNN-fc8 & {\bf 95.37} & 48.77 \\ \midrule
SIFT$_{(k_{\text{SIFT}})}$ & $78.70_{(20)}$ & $53.70_{(20)}$ \\
Pixel & 45.37 & 42.90 \\ \midrule
Random & \multicolumn{2}{c}{80.09} \\ \bottomrule
\end{tabular}
\end{table}

Finally, we experimented on the STONEFLY9 data set. As
mentioned in the data set description, this data set is difficult and
the classes are fine-grained. CaffeNet's most common predictions for
these images are mixture of sky (\texttt{kite}, \texttt{wing},
\texttt{warplane}) 
and ocean (\texttt{jellyfish}) classes that contain isolated objects
on a bright blue background. 
This experiment tests whether CaffeNet can distinguish between highly
similar classes in a different non-ImageNet domain. 

Figure~\ref{fig:exp-s9} and Table~\ref{tab:s9-auc} show  that random
selection achieved a high discovery performance, due to the small
number of classes and mostly balanced class distribution. Once again,
DEMUD was the highest performing method and the only one to
out-perform the random baseline, with fc8 yielding the best result.

\subsection{Visualization of Explanations}
\label{sec:viz-results}

As previously discussed, novelty detection methods are most useful
when they are accompanied by explanations that highlight the novel
component of a given selection.  DEMUD-VIS provides explanations
that consist of the reconstruction ($\hat \mathbf{x}'$) and the
residual ($\mathbf{r}$) feature vectors. In this section, we examine
visual explanations generated by different CNN layers and examples
of two kinds of novel discovery: new classes and variation within
classes.
All visualizations for experiments presented are 
available at \url{https://jakehlee.github.io/visualize-img-disc.html}. 

\subsubsection{Comparison of Explanations from Different CNN Layers}

\begin{figure}[]
  \centering
  \subfigure[Figure annotations]{
    \includegraphics[width=0.475\textwidth]{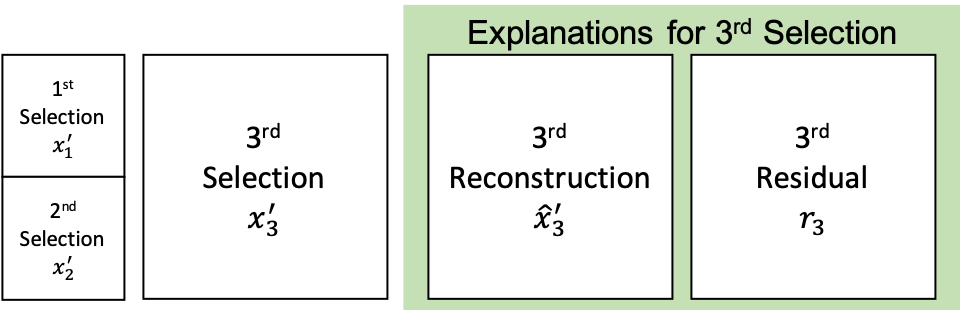}
  }
  \subfigure[CaffeNet fc6 (\texttt{school bus})]{
    \includegraphics[width=0.475\textwidth]{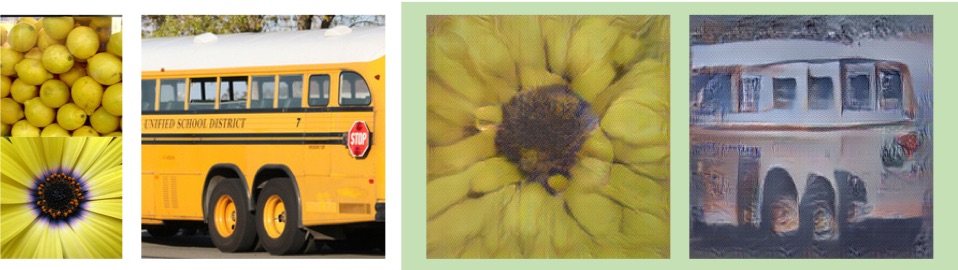}
  }
  \subfigure[CaffeNet fc7 (\texttt{daisy})]{
    \includegraphics[width=0.475\textwidth]{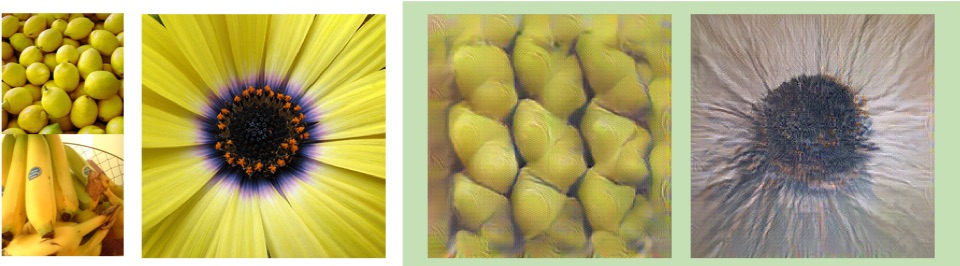}
  }
  \subfigure[CaffeNet fc8 (\texttt{rapeseed})]{
    \includegraphics[width=0.475\textwidth]{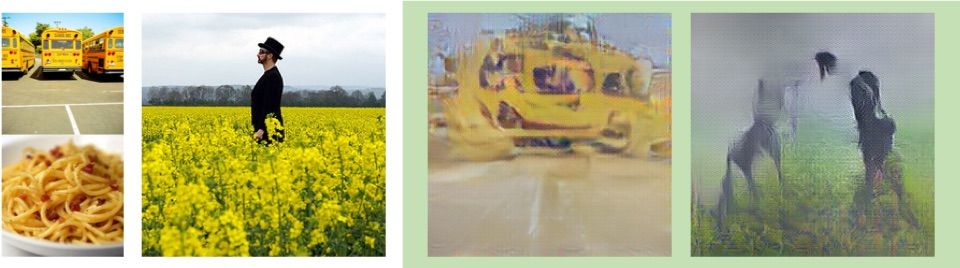}
  }
  \caption{Explanations for DEMUD-VIS selections from ImageNet-Yellow using
  CaffeNet fc6, fc7, or fc8 representations.  For each subfigure, the
  first three selected images are shown on the left. The right two
  images (green background) are the explanations generated for the
  third selection.  The \edit{}{visualized} reconstruction leverages content from the
  first two selections, while the residual shows novel content present
  only the third selection.} 
  \label{fig:layer-vis}
\end{figure}

We compared novelty detection using feature vectors extracted from the
fc6, fc7, and fc8 layers of CaffeNet, which encode increasingly abstract
image content.  
Different layers yield different selections and
explanations. Figure~\ref{fig:layer-vis} compares explanations for 
DEMUD-VIS selections from the ImageNet-Yellow data set when using
each of these layers.  For each layer experiment, we show DEMUD-VIS's
third selection ($x'_3$), which is the most novel image given the
previous two selections ($x'_1$ and $x'_2$).  The explanation for
novelty in $x'_3$ (green background) includes the
\edit{}{visualization of the} reconstruction 
$\hat{x}'_3$ and the residual $r_3$. 
The \edit{}{visualized} reconstruction contains content that DEMUD-VIS was able to model based
on the first two images, while the residual contains the novel
content.
Note that, in general, the \edit{}{visualized} reconstructions shown here are expected to
be poor matches for their respective selected images, since DEMUD-VIS
at each round selects the item that is most poorly modeled.  Images
for which the reconstruction is very good are (by definition) not
considered to be novel. 

The fc6 explanation is clear and easily interpretable. The
\edit{}{visualized} reconstruction combines content from the first two images
(\texttt{lemon} and \texttt{daisy}) that approximates the color of the
\texttt{school bus} but fails to reproduce its novel components
(including windows, wheels, stop sign), as shown in the residual.  The
residual appears washed out because the vivid yellow tones were
accurately captured in the \edit{reconstruction}{visualization}.  The large discrepancy
between image and reconstruction caused this image to be selected as
novel.
%
Similarly, the fc7 explanation is easy to interpret.  The
\edit{}{visualized} reconstruction is mostly composed of lemons (the best color match),
while the residual emphasizes the dark center and the radial texture
of the flower petals.  
%
However, the fc8 explanation is less interpretable. The reconstruction
visualization employs content from the \texttt{school bus} image, but 
it is highly deformed. The residual is blurred and inaccurate compared
to the selected image.

This pattern occurs with other data sets and aligns with our
understanding of neural network abstraction: fc6 is closer to the
original image content, while fc8 is closer to abstract class-level
content.  It is expected that UC DeePSiM visualization quality
degrades as the feature space moves away from the convolutional layers
(conv5) and closer to the final softmax classification
layer~\citep{dosovitskiy:gen16,dosovitskiy:upconv16}. 
We are presented with a trade-off between explanation interpretability
and class discovery performance. DEMUD-VIS with fc8 features achieved the
best discovery performance across all data sets
(Section~\ref{sec:disc-perf}), but it did not generate consistently
interpretable explanations. DEMUD-VIS with fc6 features offers discovery
performance that out-performs most other methods and yields
the most interpretable explanations. Therefore, all visualizations 
in this subsection will be generated from fc6 features. 

\subsubsection{Explanations of First-Time Class Discovery}

\begin{figure}
  \centering
  \subfigure[4$^\text{th}$ Selection]{
    \includegraphics[width=0.22\textwidth]{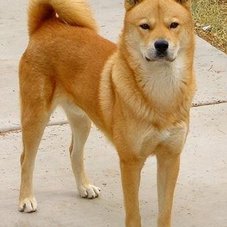}
  }
  \subfigure[Reconstruction]{
    \includegraphics[width=0.22\textwidth]{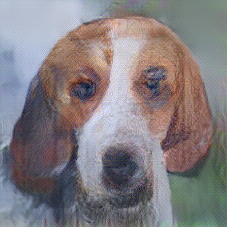}
  }
  \subfigure[Residual]{
    \includegraphics[width=0.22\textwidth]{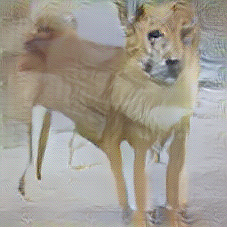}
  }
  \caption{Explanation for the DEMUD-VIS discovery of the \texttt{Eskimo dog,
    husky} class (selection 4) in the balanced ImageNet-Random data set
    . Previous selections included images from the \texttt{English
    foxhound}, \texttt{projectile, missile}, and \texttt{baseball}
    classes.}
  \label{fig:discvis-random}
\end{figure}

\begin{figure}
  \centering
  \subfigure[6$^\text{th}$ Selection]{
    \includegraphics[width=0.22\textwidth]{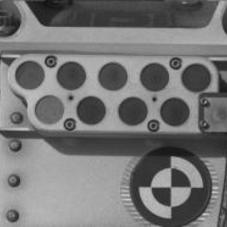}
  }
  \subfigure[Reconstruction]{
    \includegraphics[width=0.22\textwidth]{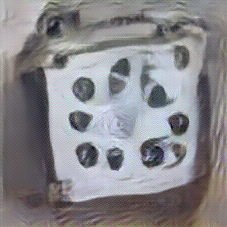}
  }
  \subfigure[Residual]{
    \includegraphics[width=0.22\textwidth]{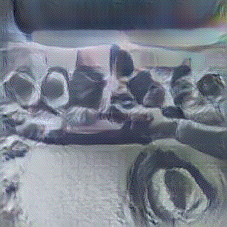}
  }
  \caption{Explanation for the DEMUD-VIS discovery of the \texttt{ChemCam
    Calibration Target} class (selection 6) in the Mars-Curiosity data
    set.  Previous selections included five distinct classes, of which
    the best match was the \texttt{REMS UV Sensor} instrument, as shown
    in the \edit{}{visualized} reconstruction.}
  \label{fig:discvis-msl}
\end{figure}

\begin{figure}
  \centering
  \subfigure[7$^\text{th}$ Selection]{
    \includegraphics[width=0.22\textwidth]{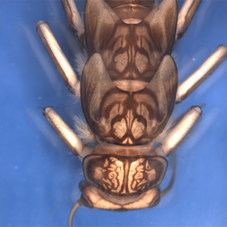}
  }
  \subfigure[Reconstruction]{
    \includegraphics[width=0.22\textwidth]{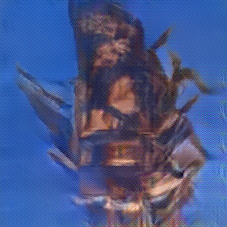}
  }
  \subfigure[Residual]{
    \includegraphics[width=0.22\textwidth]{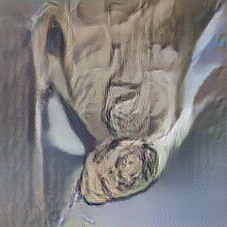}
  }
  \caption{Explanation for the DEMUD-VIS discovery of the \texttt{CAL} class
    (selection 7) in the STONEFLY9 data set. Previous selections
    included six images from four different taxa, none of which had the
    light-colored markings and texture of this specimen.}
  \label{fig:discvis-s9}
\end{figure}

When DEMUD-VIS discovers a new class, the explanation helps pinpoint how
it is new.  Figure~\ref{fig:discvis-random} shows the DEMUD-VIS explanation
when it discovered the \texttt{Eskimo dog, husky} class in the balanced
ImageNet-Random data set.  The model had previously selected an image
from the \texttt{English foxhound} class, which dominates the
\edit{}{visualized} reconstruction. The residual emphasizes differences the in coat color
and pose in the new image.

Figure~\ref{fig:discvis-msl} shows another example of class discovery,
this time from the Mars-Curiosity data set.  This selection constitutes
the discovery of the \texttt{ChemCam Calibration Target} class. The
model had previously selected five images from five distinct classes,
one of which contained the \texttt{REMS UV Sensor}. This previously
selected image contains small, dark circles on a white background,
similar to the \texttt{ChemCam Calibration Target} image. However, the
\texttt{ChemCam Calibration Target} image differs in the location and
pattern of the dark circles, and also contains a larger patterned
circle on the bottom right. These features are highlighted in the
residual visualization.

Finally, Figure~\ref{fig:discvis-s9} shows an example of the discovery
of the \texttt{CAL} class in the STONEFLY9 data set. The model had
previously selected six images from four different classes (taxa), one
of which was a dark insect in a similar orientation. The
\edit{}{visualized} reconstruction
replicates the dark insect, while the residual highlights the
light-colored markings in the selected image. Note that the blue
background is not present in the residual, because it was learned from
the previous selections and therefore is not novel.

In each case, the explanations illustrate image content that can be
modeled (reconstructed) from previous selections as well as content
that is new, which helps humans realize that a new class is present
and provides the first step in characterizing the new class.

\begin{figure}
  \centering
  \subfigure[15$^\text{th}$ selection]{
    \includegraphics[width=0.22\textwidth]{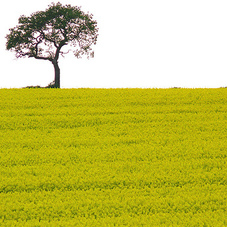}
  }
  \subfigure[Reconstruction]{
    \includegraphics[width=0.22\textwidth]{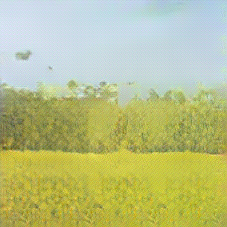}
  }
  \subfigure[Residual]{
    \includegraphics[width=0.22\textwidth]{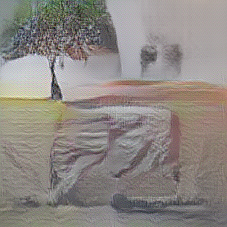}
  }
  \caption{\edit{}{Explanation for the similarity-based retrieval of
  the
  \texttt{rapeseed} class
    (selection 15) in the ImageNet-Yellow data set. Previous selections
    included 14 images from four different classes, half of which
    were in the \texttt{rapeseed} class.}}
  \label{fig:simvis}
\end{figure}

\edit{}{At this stage, it is worth noting that the examples in
  Figures~\ref{fig:discvis-random}-\ref{fig:discvis-s9} often include
visualized reconstructions that do not match the original
selected images. This is not due to a problem in the visualization method,
but because DEMUD-VIS deliberately selects the image with the greatest
reconstruction error (in feature space), which corresponds to
maximizing distance from what has been previously seen.} 

\edit{}{To show the visualization method's capabilities, we ran a
similarity-based retrieval experiment with the ImageNet-Yellow dataset
that
instead selects the item with the {\em lowest} reconstruction
error. An example 
from this experiment is shown in Figure~\ref{fig:simvis}. Previous
selections included many examples of the \texttt{rapeseed} (yellow
flower) class, but no solitary trees.  DEMUD-VIS therefore reconstructed all
parts of the selected image except for the tree on the top left. The
same tree is the only feature that remains in the visualization of the
residual.} 

\subsubsection{Explanations of Novelty Discovery Within a Class}

\begin{figure}
  \centering
  \subfigure[2$^\text{nd}$ Selection]{
    \includegraphics[width=0.22\textwidth]{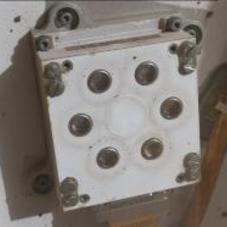}
  }
  \subfigure[15$^\text{th}$ Selection]{
    \includegraphics[width=0.22\textwidth]{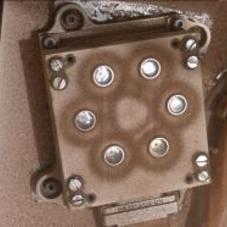}
  }
  \subfigure[Reconstruction]{
    \includegraphics[width=0.22\textwidth]{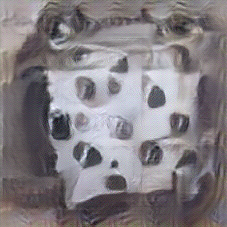}
  }
  \subfigure[Residual]{
    \includegraphics[width=0.22\textwidth]{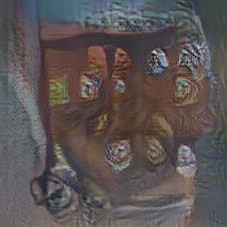}
  }
  \caption{DEMUD-VIS selected two images from the \texttt{REMS UV Sensor} class
    due to changes in appearance caused by dust deposition.
  The first image (a) was taken on sol 36, and the second image (b) was taken
  on sol 808.}
  \label{fig:msl-time}
\end{figure}

Novelty detection is not restricted to the discovery of new classes.
Figure~\ref{fig:msl-time} shows an example two very different images
from the same class (\texttt{REMS UV Sensor}) that DEMUD-VIS selected from
the Mars-Curiosity data set. While the second image does not constitute
a new class discovery, the sensor's appearance has significantly
changed to due dust deposition in the 772 sols (Mars days) that elapsed
between the two images.  The explanation shows that the model attempted
to reconstruct the sensor in its clean state, and the residual
emphasizes the darker, dirtier appearance in the new image.

\begin{figure}
  \centering
  \subfigure[1$^\text{st}$ Selection]{
    \includegraphics[width=0.22\textwidth]{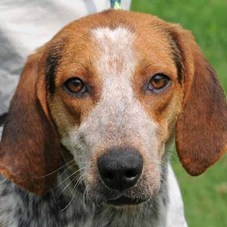}
  }
  \subfigure[35$^\text{th}$ Selection]{
    \includegraphics[width=0.22\textwidth]{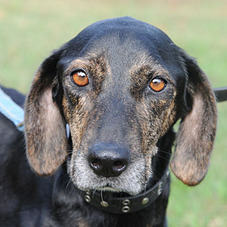}
  }
  \subfigure[Reconstruction]{
    \includegraphics[width=0.22\textwidth]{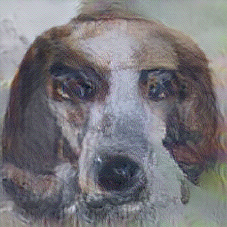}
  }
  \subfigure[Residual]{
    \includegraphics[width=0.22\textwidth]{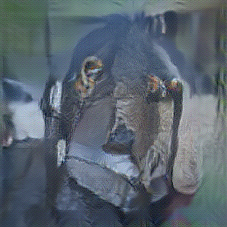}
  }
  \caption{Two selections from the \texttt{English foxhound} class with
    very different coat coloring.}
  \label{fig:foxhounds}
\end{figure}

Figure~\ref{fig:foxhounds} shows another example of within-class
novelty from the imbalanced ImageNet-Random data set.  Both images are
from the \texttt{English foxhound} class with similar poses.  However,
as highlighted in the explanation, they have very different coloring
(brown and white versus black).  Whether such differences constitute
class-level novelty depends on the granularity of interest to the user,
and being made aware of this diversity can help guide those decisions.

\subsection{User Study to Assess Explanation Utility}
\label{sec:userstudy}

Thus far, the novelty explanations displayed have been anecdotal and
interpreted by the authors.  To independently assess the utility of these
explanations for the general public, we conducted a large user study.
We presented users with 20 images and asked them to rate the novelty
of each one.  We employed a randomized A/B testing strategy in which
some users were given only the image and others were given the image
plus the visual novelty explanation.
\paragraph{User study design.}
Users were shown the first 20 images selected by DEMUD-VIS from the
ImageNet-Yellow data set, as described in Section~\ref{sec:datasets}. Images
were presented in the order of their selection (most novel first).  We
chose to use 20 images chosen from a data set of 10 classes to ensure
that there would be a mixture of novel 
and not-novel images, with respect to their ImageNet class labels. As
discussed in the previous section, even within a given class there
can be images with unique or novel elements.


For each selection, users were asked to answer the question
``Given the images you have already seen, does the next image
contain something new?'' The response options provided were:
\begin{itemize}
\item Yes: new type of object
\item Some: new aspect of object type
\item Minor: unimportant details are new
\item No: duplicate of previous object
\item Not sure / skip
\end{itemize}

We proposed two hypotheses to assess the utility of the explanations:
\begin{enumerate}
\item Explanations increase user confidence: 
  Users will have fewer ``Not sure'' responses when given
  the explanation, compared to no explanation provided.
\item Explanations increase user awareness of novelty:
  Users will have more ``Yes'' responses when given the explanation, 
  compared to no explanation.
\end{enumerate}

\paragraph{Response characterization.}
We used the SurveyMonkey platform to collect responses over a period
of two months.  
We advertised the survey online, within our
social networks, at Columbia University, and at the Jet Propulsion
Laboratory.  We received 399 responses, of which 280 (70\%) were
complete.  The response rate was higher for survey A (without
explanations) (80\%) than for survey B (with explanations) (61\%).
This suggests that using the explanations required more effort, 
motivation, or persistence on the part of the participant. 
The average time to complete the survey was 366 seconds for survey A
(standard deviation of 888.9 s)
and 457 seconds for survey B (standard deviation of 262.7
s)\footnote{We excluded one survey B 
  response from our analysis that had a duration of 1,437,922 seconds
  (16 days) on the grounds that this likely did not represent
  continuous effort.  Since the survey assesses novelty, it is
  important that it is completed in one session.}, 
but this difference was not significant at the 95\% confidence level
(using Welch's unequal variances t-test). 

\begin{figure}
\centering
\subfigure[Education]{\includegraphics[width=2.3in]{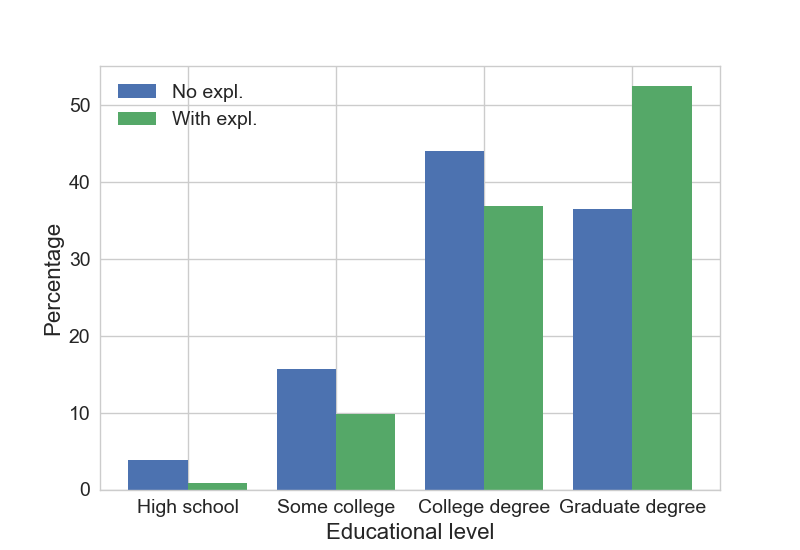}}
\subfigure[ML experience]{\includegraphics[width=2.3in]{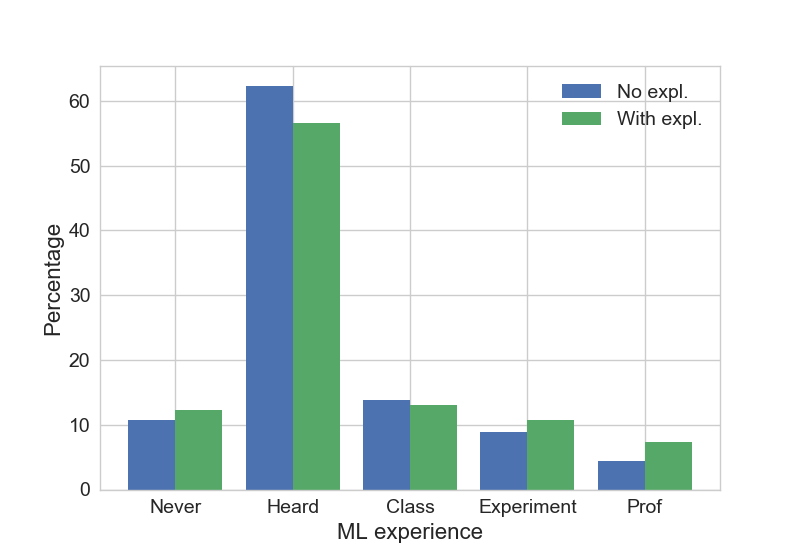}}
\caption{Demographics of user study participants who completed the survey.}
\label{fig:demog}
\end{figure}

We also found a difference in the distribution of educational
backgrounds for those who completed each survey, despite random
assignment to survey A or B.  Figure~\ref{fig:demog} shows that those
who completed survey B (with explanations) tended to have higher educational
levels than did survey A completers.  Less difference was observed in
the amount of self-reported experience with machine learning, although
survey B completers were slightly more likely to have run their own
machine learning experiments or worked as professionals in the field.

\paragraph{Results.}
We compiled the results of the 280 complete responses to test the
hypotheses stated above.

{\bf Hypothesis 1:} We found a significant overall difference ($p=$
\SI{3e-5}, using the Fisher exact test) that indicates survey B
respondents were in general 4 times {\em more} likely to respond ``Not
sure'' than survey A respondents. 
This contradicts our hypothesis that survey B respondents would be
less likely to respond ``Not sure.''  However, this occurred only 9 of
3180 times for survey A and 30 of 2420 times for survey B.
The rarity of this response (only 0.7\% of responses) suggests that
users generally felt sufficiently confident that they could rate the
novelty of a given image regardless of whether the explanation was
provided.  For a data set in which novelty manifests in more subtle
ways, the outcome may be different.
We did not find a statistically significant difference in the number
of ``Not sure'' responses to the two surveys for any individual images.  

\begin{figure}
\centering
\subfigure[ImageNet class: \texttt{corn}]
{\includegraphics[width=4in]{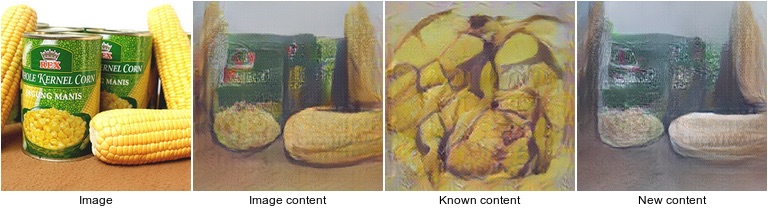}}
\subfigure[ImageNet class: \texttt{daisy}]
{\includegraphics[width=4in]{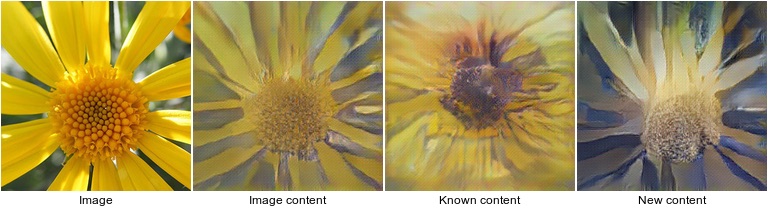}}
\subfigure[ImageNet class: \texttt{banana}]
{\includegraphics[width=4in]{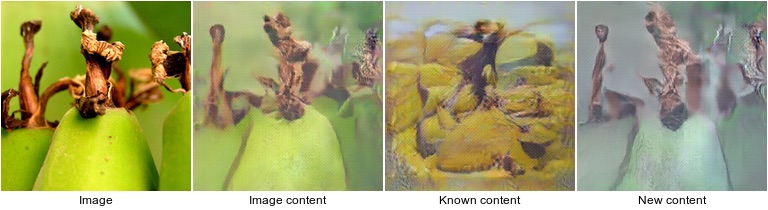}}
\caption{Images more likely to be voted new when a novelty explanation
  is provided; each one shows within-class novelty.}
\label{fig:Bnew}
\end{figure}

{\bf Hypothesis 2:} 
Overall, survey B users were slightly more likely to respond ``Yes''
(image contains a new type of object)
than were survey A users, but the difference was not significant.
However, there are differences in which images they considered to be
novel, with survey B users showing more sensitivity to within-class novelty.
%
For three of the 20 images (shown in Figure~\ref{fig:Bnew}), survey B
users were more likely than 
survey A users to confidently state that the image contained a new
type of object ($p \leq 0.05$), consistent with our hypothesis.
Note that to make the explanations more widely accessible, we called
the reconstruction ``known content'' and the residual ``new content.''
Each of these images were cases of within-class novelty, since the
same class had appeared earlier in the selections seen by the user.
As shown in Figure~\ref{fig:Bnew}, a \texttt{corn} image includes a green
can; a \texttt{daisy} image includes a yellow (not brown) center, and a
\texttt{banana} image is dominated by the stem and very green bananas.
Within-class novelty is often more subtle than the discovery of a new class.
Survey B users, who received the novelty explanation highlighting
these differences, judged the images to be novel more frequently than
did Survey A users, who did not have the aid of the explanation.  

\begin{figure}
\centering
\subfigure[ImageNet class: \texttt{daisy} (previous: \texttt{school
    bus} and \texttt{lemon})] 
{\includegraphics[width=4in]{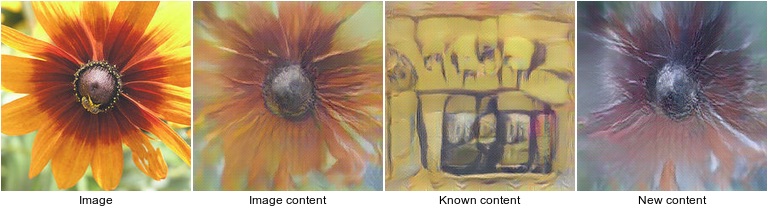}}
\subfigure[ImageNet class: \texttt{corn} (previous: \texttt{school
    bus}, \texttt{lemon}, and \texttt{daisy})] 
{\includegraphics[width=4in]{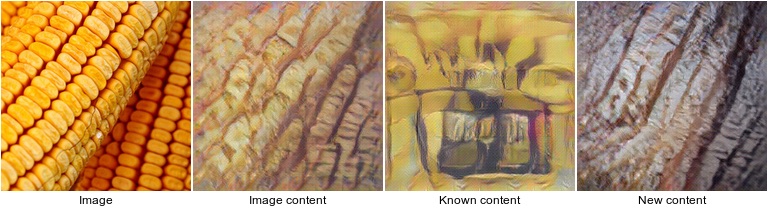}}
\subfigure[ImageNet class: \texttt{spaghetti squash} (previous:
  \texttt{school bus}, \texttt{lemon}, \texttt{daisy}, \texttt{corn},
  and \texttt{lemon})] 
{\includegraphics[width=4in]{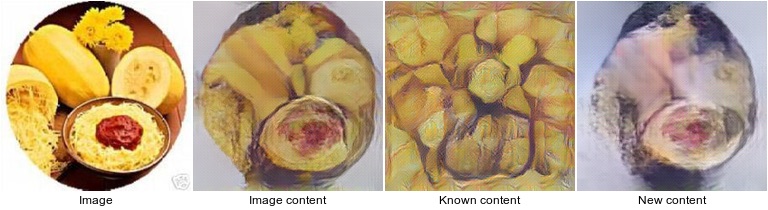}}
\caption{Images {\em less} likely to be voted new when a novelty explanation
  is provided.}
\label{fig:Bnotnew}
\end{figure}

In contrast, for three other images (see Figure~\ref{fig:Bnotnew}),
survey A users were more likely 
than survey B users to vote ``Yes'' for novelty.  These three images
were all examples of ImageNet classes the
user had not yet seen, so they were genuinely new.  There is no clear
explanation for why users given explanations would be {\em less} likely to
consider these images novel.  They all occurred early in the sequence
(selections 2, 3, and 5), so we speculate that users were confused
by or unable to interpret the explanations at this point in the
survey.  As they gained more experience, they were able to extract
more meaning from the explanations, leading to the more fine-grained decisions
that were made in Figure~\ref{fig:Bnew}, which occurred in selections 14,
16, and 17.  This observation provides useful feedback that additional
user training prior to working with explanations, or the ability to
reference the tutorial while browsing explanations, could benefit
future users. 







\section{Conclusions and Future Work}
\label{sec:conc}

We have developed and evaluated the first approach to generating
visual explanations of novel image content.  This technique can
provide a prioritized list of unusual or anomalous images within a
large data set to enable the best use of limited human review time.
Images with novel content may signal the occurrence of unanticipated
artifacts (e.g., data corruption) or the potential for a new
scientific discovery (e.g., a new insect species).
Our approach combines the use of a convolutional neural network (CNN)
to extract features to represent image content, a novelty detection
method based on reconstruction error, and an up-convolutional network
to convert novelty explanations from the CNN feature space back into
image space for human interpretation.

This approach achieves strong (sometimes near-perfect) performance in
discovering new classes within large data sets for a range of image
domains, including standard ImageNet images, microscope images of
insects collected in freshwater streams, and images collected by a
rover on Mars.
We evaluated the utility of the corresponding image explanations
through a user study and found that visual novelty explanations
influenced user decisions about novelty by highlighting specific
aspects of the image that were new.  The explanations increased the
chance that users detected within-class novelty.  We observe that
there is a human learning curve, and we expect that users would
benefit from seeing multiple example explanations so that they are
well positioned to understand the explanations.  

\edit{}{\textbf{Limitations:} Our method is reliant on the properties
of the CNN used for feature extraction. Novelty detection can only be
performed in context of features that the CNN is able to extract. For
example, if the image features extracted by the CNN are invariant to
image translation, but the domain considers feature location
significant, then our method will miss novel images in this context.}

\edit{}{Second, because this approach is unsupervised, it cannot
  determine when all classes of interest have been found during a
  class discovery task. 
DEMUD makes no distinction between inter-class and intra-class
novelty; an item selected as novel may represent a newly discovered
class or diversity within a class. User interpretation is needed to
determine whether a change
in color, pose, texture, etc. signifies a new class for a given domain.
The user can specify a limit on the number of selections to
be generated, or the method can be allowed to exhaust the data set,
resulting in a complete ranking.  In either case, selections will be
provided in decreasing order of novelty.}

\edit{}{Finally, the user study demonstrated that there is still room
for improvement in the quality of the generated explanations. The
current study used CaffeNet and its corresponding visualization
network to take advantage of pre-trained networks, but using a more
advanced network such as ResNet-50 \citep{he:resnet16} for both
feature extraction and visualization may result in better
performance.
Visualization of the residual (what is new) is more challenging.
Since reconstruction error is maximized for each selection, in the
first few selections we often find that effectively everything about
the image is novel and the residual visualization shows almost the
same content as the original image.  In later selections, the novelty
may be more subtle or abstract, such as a change in pose or time of
day, which can be difficult to visualize independent of the object
that is present.  In contrast, we have found that changes in color or
texture are expressed well.  It is an open area of study to identify
the kinds of novelty that do or do not lend themselves well to visual
representation. 
}

\edit{}{\textbf{Applications:}} This approach can be used to 
accelerate analysis and discovery in a
variety of application areas, ranging from surveillance to remote
sensing to medical
diagnosis~\citep{chandola:anom09,pimental:novelty14}.  Many 
investigations employ cameras to observe phenomena of interest.  By
focusing attention on the most novel images, our approach can help
investigators quickly zero in on the observations most likely to lead
to new discoveries.

\begin{figure}[b]
  \centering
  \subfigure[Selection]{
    \includegraphics[width=0.25\textwidth]{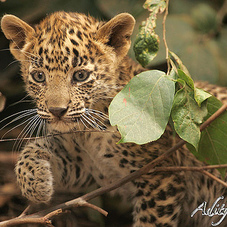}
  }
  \subfigure[Reconstruction]{
    \includegraphics[width=0.25\textwidth]
    {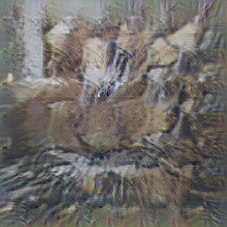}
  }
  \subfigure[Residual]{
    \includegraphics[width=0.25\textwidth]
    {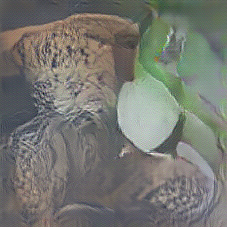}
  }
  \caption{DEMUD-VIS discovered this mislabeled ImageNet image: the
    leopard cub is mislabeled as a tiger cub.}
  \label{fig:tigerleopard}
\end{figure}

We have also found that DEMUD-VIS is very good at detecting mislabeled 
examples, since they contain unusual content with respect to their
class. In a separate experiment with ImageNet data, we applied
DEMUD-VIS to 63 images from the \texttt{tiger cub} class.  The ninth
selection turned out to be a leopard cub that does not belong to this
class.  As shown in Figure~\ref{fig:tigerleopard}, the reconstruction
is very poor, because DEMUD-VIS had only seen tiger cubs and could not
reproduce the leopard.  Tiger-stripe-like features are evident (and
incorrect).  The residual highlights the green leaf and the leopard's
spots, which are novel and help diagnose the ImageNet labeling error.
This capability could be useful in exploring even fully labeled data
sets, to help identify labeling errors and/or adversarial examples.

\begin{acknowledgements}
We thank the Planetary Data System Imaging Node for funding this
project. Part of this research was carried out at the Jet Propulsion
Laboratory, California Institute of Technology, under a contract
with the National Aeronautics and Space Administration.  All rights
reserved. \copyright 2019 California Institute of
Technology. U.S. Government sponsorship acknowledged. 
\end{acknowledgements}

\bibliographystyle{spbasic}      
\bibliography{refs}   

\end{document}